\documentclass[10pt,twocolumn]{article} 
\usepackage{simpleConference}
\usepackage{times}
\usepackage{graphicx}
\usepackage{amssymb}
\usepackage{url,hyperref}
\usepackage{amsmath}
\usepackage{caption}
\usepackage{multirow}

\begin{document}

\title{PolarDet: A Fast, More Precise Detector for Rotated Target in Aerial Images}

\author{Pengbo Zhao\textsuperscript{\rm 1,2},
	Zhenshen Qu\textsuperscript{\rm 1,*},
	Yingjia Bu\textsuperscript{\rm 2},
	Wenming Tan\textsuperscript{\rm 2},
	Ye Ren\textsuperscript{\rm 2},
	Shiliang Pu\textsuperscript{\rm 2} \\
	\textsuperscript{\rm 1}Harbin Institute of Technology \\
	\textsuperscript{\rm 2}Hikvision Research Institute \\
	19S004031@stu.hit.edu.cn, miraland@hit.edu.cn, \{Yingjia Bu,Wenming Tan,Ye Ren,Shiliang Pu\}@hikvision.com
}

\maketitle
\thispagestyle{empty}

\begin{abstract}
Fast and precise object detection for high-resolution aerial images has been a challenging task over the years. Due to the sharp variations on object scale, rotation, and aspect ratio, most existing methods are inefficient and imprecise. In this paper, we represent the oriented objects by polar method in polar coordinate and propose PolarDet, a fast and accurate one-stage object detector based on that representation. Our detector introduces a sub-pixel center semantic structure to further improve classifying veracity. PolarDet achieves nearly all SOTA performance in aerial object detection tasks with faster inference speed. In detail, our approach obtains the SOTA results on DOTA, UCAS-AOD, HRSC with 76.64\% mAP, 97.01\% mAP, and 90.46\% mAP respectively. Most noticeably, our PolarDet gets the best performance and reaches the fastest speed(32fps) at the UCAS-AOD dataset.
\end{abstract}

\section{Introduction}
\begin{figure}[t]
	\centering
	\includegraphics[width=1.0\linewidth]{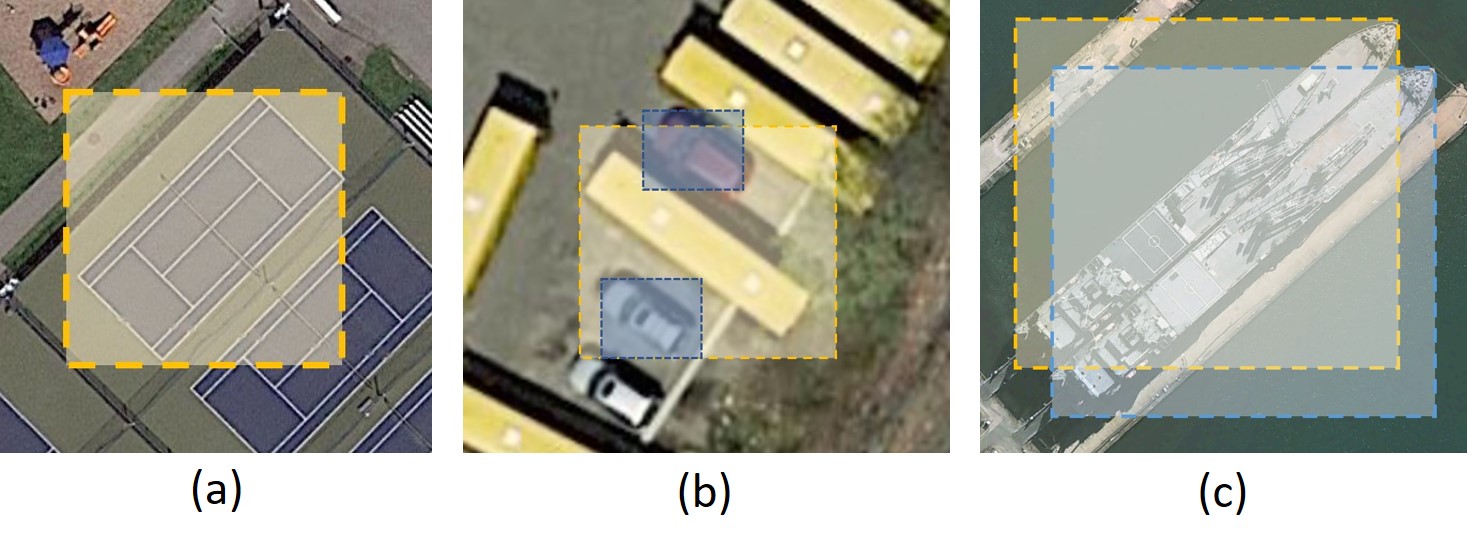}	
	\caption{Existing shortages in current methods, (a)incorrect cover(same category), (b)incorrect cover(different category), (c)wrong suppression because of NMS}
	\label{fig:existing problem}
\end{figure}

Recently, object detectors \cite{uijlings2013selective}, \cite{girshick2014rich}, \cite{he2015spatial}, \cite{girshick2015fast}, \cite{ren2015faster}, \cite{dai2016r}, \cite{redmon2016you} based on convolutional neural network (CNN) have got many achievements in nature scene detection. Yet many shortages still exist when these methods straight transferred to aerial image. As shown in Fig.\ref{fig:existing problem}, the horizontal bounding box, which is oversized, will cover the background and create wrong suppression. To avoid these problems, many detectors \cite{yang2019r3det}, \cite{azimi2018towards}, \cite{ding2019learning} use five-parameter method with $\theta$ $(x,y,w,h,\theta)$ to express object orientation. However, due to the sharp change in rotation, one angle expression will cause many defects, such as precision decrease, angle boundary missing, and angle loss trap (as discussed in \ref{polar_angle}). Also, directly regressing $w$ and $h$ will drop the network convergence performance. That is because of sharp variations on the object scale and aspect ratio. To increase the expression precision and avoid angle loss trap, eight-parameter method $(\delta x_i,\delta y_i,(i=1,2,3,4))$ is proposed by some detectors. But these methods still cannot resolve the decrease in network convergence performance.

\begin{figure}[t]
	\centering
	\includegraphics[width=1.0\linewidth]{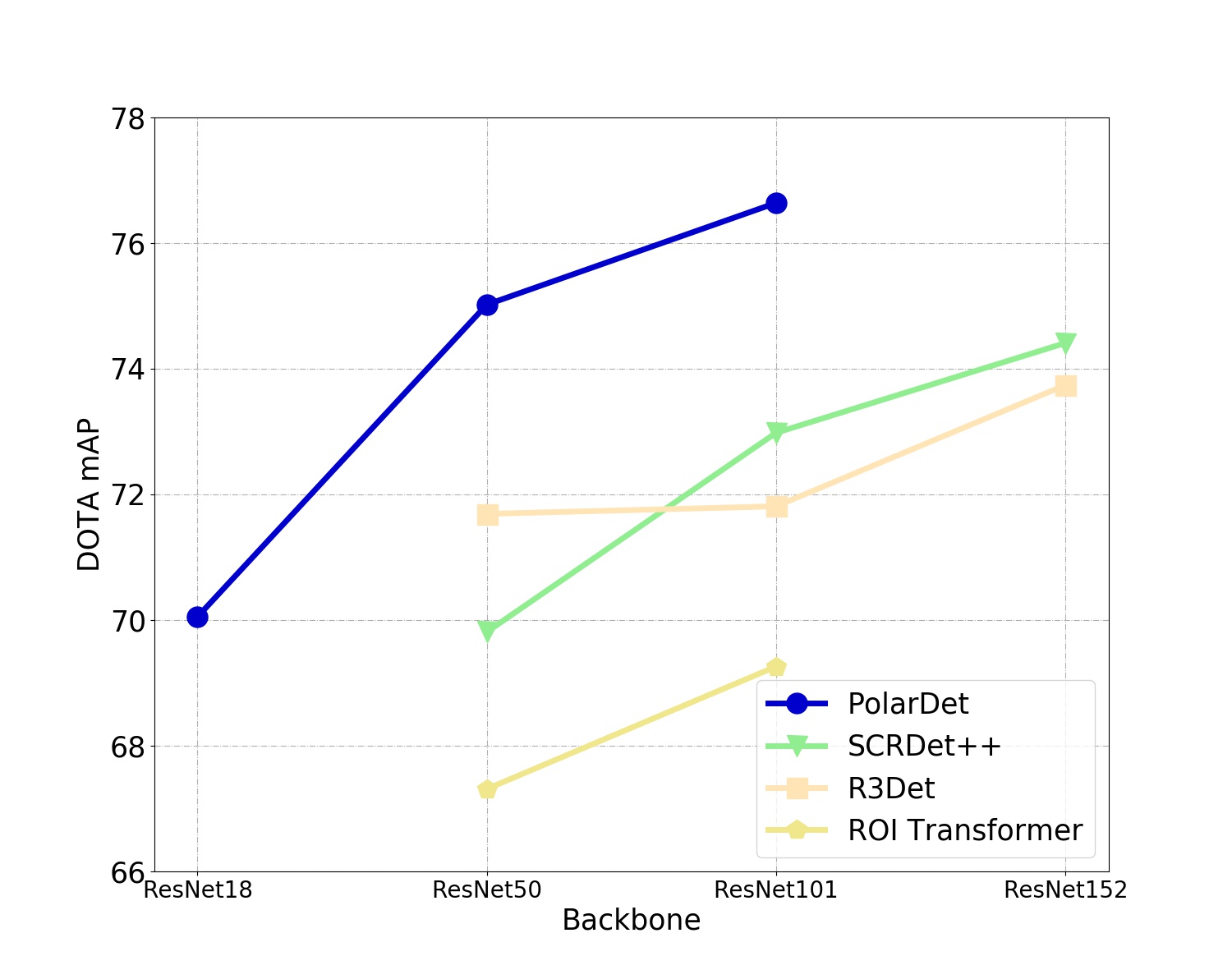}	
	\caption{mAP of different approaches of DOTA dataset. The proposed PolarDet outperforms nearly all the other algorithm and gets SOTA results.}
	\label{fig:SOTA results}
\end{figure}

\begin{figure}[t]
	\centering
	\includegraphics[width=1.0\linewidth]{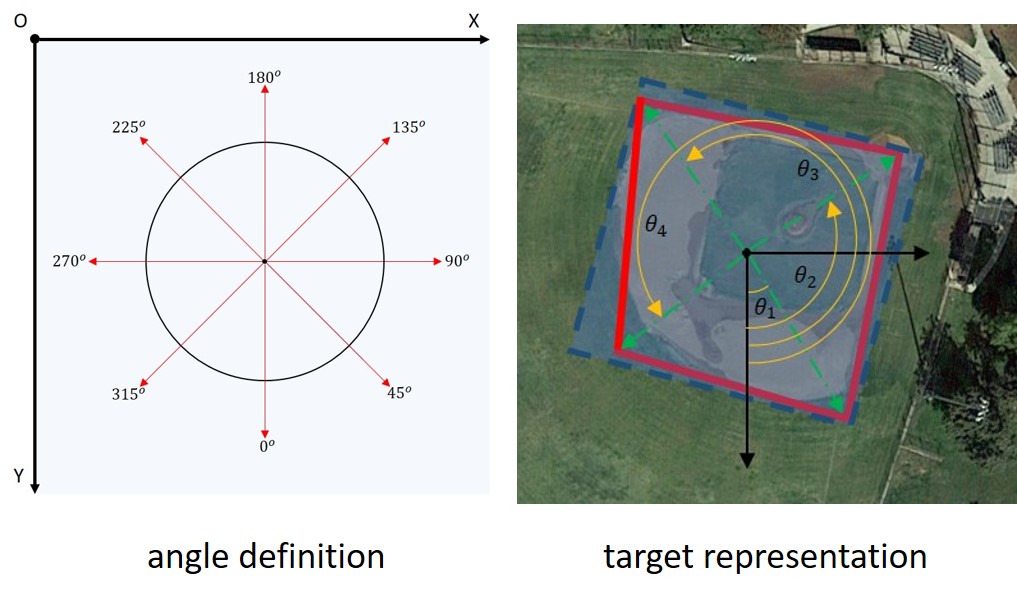}	
	\caption{Based on polar coordinate, polar method represents target by $((x,y),(\delta x,\delta y),\theta_{p},s,r_{p})$}
	\label{fig:coordinate representation}
\end{figure}

In this paper, we propose a fast and more precise alternative polar method, called PolarDet. Based on the polar coordinate, we represent the target by multiple angles and shorter-polar diameter ratio. To be detailed, as shown in Fig.\ref{fig:coordinate representation}, we represent target by $((x,y),(\delta x,\delta y),\theta_{p},s,r_{p},(p=1,2,3,4))$. Concretely, targets will be described by a center point, offset, four polar-angle, one shorter side, and four polar-ratio. They represent the center of the target, the offset of the target center, the angles between four polar diameters and the reference y-axis (will be explained in Section \ref{TPM}), the shorter one between the minimum bounding rectangle width and its height, the ratio between shorter side and polar diameter, respectively. The polar diameter here represents the Euclidean distance between the center point and corner. With the four angles prediction, our polar method can express orientation more precisely and avoid angle loss trap. With the relative polar diameter regression (shorter side and polar-ratio), our polar method also can increase network convergence performance. We will discuss how it works and its advancement in Section \ref{TPM}.


Besides the polar method, we also introduce an improved center semantic Structure to enhance the classification capability of our network. Most importantly, our result is got from the single ResNet \cite{he2016deep} network as backbone and without FPN structure. With rarely extra parameters increasing, this achievement satisfies the need for speed and meets the requirement of accuracy at the same time.

This paper makes the following contributions:

(1) We propose a fast and more precise detector PolarDet, where we represent the targets by $((x,y),(\delta x,\delta y),\theta_{p},s,r_{p})$. This representation can resolve most of the defects that current methods face. 

(2) We introduce an improved center semantic structure, which can enhance the precision of classification without adding much parametera.

(3) As shown in Fig.\ref{fig:SOTA results}, we achieve the SOTA results on both of the DOTA \cite{xia2018dota} dataset, UCAS-AOD \cite{li2019feature} dataset, and HRSC2016 \cite{liu2016ship} dataset. On DOTA dataset, we reach 76.64\% mAP with ResNet-101 as backbone. On AOD and HRSC datasets, we attain 97.02\% mAP and 90.46\% mAP with ResNet-50 backbone respectively.

\section{Related Work}\label{Related_Work}

We will show the detectors using horizontal methods in \ref{HOD} and show the improved oriented methods in \ref{OOD}. Also, we will list the commonly used attention mechanism which is related to our center semantic structure.

\subsection{Horizontal Object Detectors}\label{HOD}

Faster R-CNN \cite{ren2015faster} first introduces horizontal anchor into target detection. FPN \cite{lin2017feature}, Cascade R-CNN \cite{cai2018cascade}, and R-FCN \cite{dai2016r} achieve better performance based on horizontal anchor. SSD \cite{liu2016ssd}, YOLO \cite{redmon2016you}\cite{redmon2017yolo}\cite{redmon2018yolov3} improve horizontal anchor strategy and increase detection speed. CornerNet \cite{law2018cornernet}, CenterNet \cite{duan2019centernet}, and ExtremeNet \cite{zhou2019bottom} propose horizontal boundary embedding points prediction. CenterNet \cite{zhou2019objects} and FCOS \cite{tian2019fcos} regard target as point then generate horizontal bounding box.

These horizontal object detectors all face many problems shown in Fig.\ref{fig:existing problem} because the sharp variations in aerial images and oriented detection tasks.

\subsection{Oriented Object Detectors}\label{OOD}

R-RPN \cite{ma2018arbitrary} directly uses rotated anchor to detect oriented target. R2CNN \cite{jiang2017r2cnn} predicts horizontal and oriented prediction box based on horizontal anchor. ROI Transformer \cite{ding2019learning}, SCRDet \cite{yang2019scrdet},  R3Det \cite{yang2019r3det}, and SCRDet++ \cite{yang2020scrdet++} apply five-parameter method while Textbox++ \cite{liao2018textboxes++} and RSDet \cite{qian2019learning} apply eight-parameter method to represent oriented target. Gliding Vertex \cite{xu2020gliding} introduce vertex gliding to locate object.

However, these oriented object detectors still face angle boundary, angle-loss trap, regression fluctuation, and imprecise fitting problems.

\subsection{Attention Mechanism}\label{AM}

STN \cite{jaderberg2015spatial} and SENET \cite{hu2018squeeze} introduce spatial and channel attention mechanism to extract key feature respectively. DANet \cite{fu2019dual} connects spatial and channel semantic feature to distinguish target. PointRend \cite{kirillov2020pointrend} utilizes boundary information in subpixel to enhance segmentation performance. 

However, these method always add too much calculation into network by using convolution operation, which results in slow inference problem.

In this paper, we introduce the polar method to represent targets more precisely. Also, we raise improved center semantic structure to further classify and locate targets.

\section{The Proposed Method}\label{TPM}


In this section, we will expound the proposed method as the following order. First, we will explain the commonly used five-parameter and eight-parameter methods and their defects in Section \ref{PCR}. To resolve these defects, we propose PolarDet and describe its pipeline in Section \ref{structure}. In our PolarDet, we propose a more precise method that represents the target in polar coordinate by multiple angles and shorter-polar diameter ratio. In Section \ref{center point} \& \ref{polar_angle} \& \ref{polar_diameter}, we will show the representation of target by our method and how it can resolve these defects. The representation consists of center point, polar angle, and polar diameter. At last, we introduce center semantic structure in Section \ref{Center_Semantic} to increase classification performance.

\begin{figure}[!tb]	
	\centering
	\includegraphics[width=1.0\linewidth]{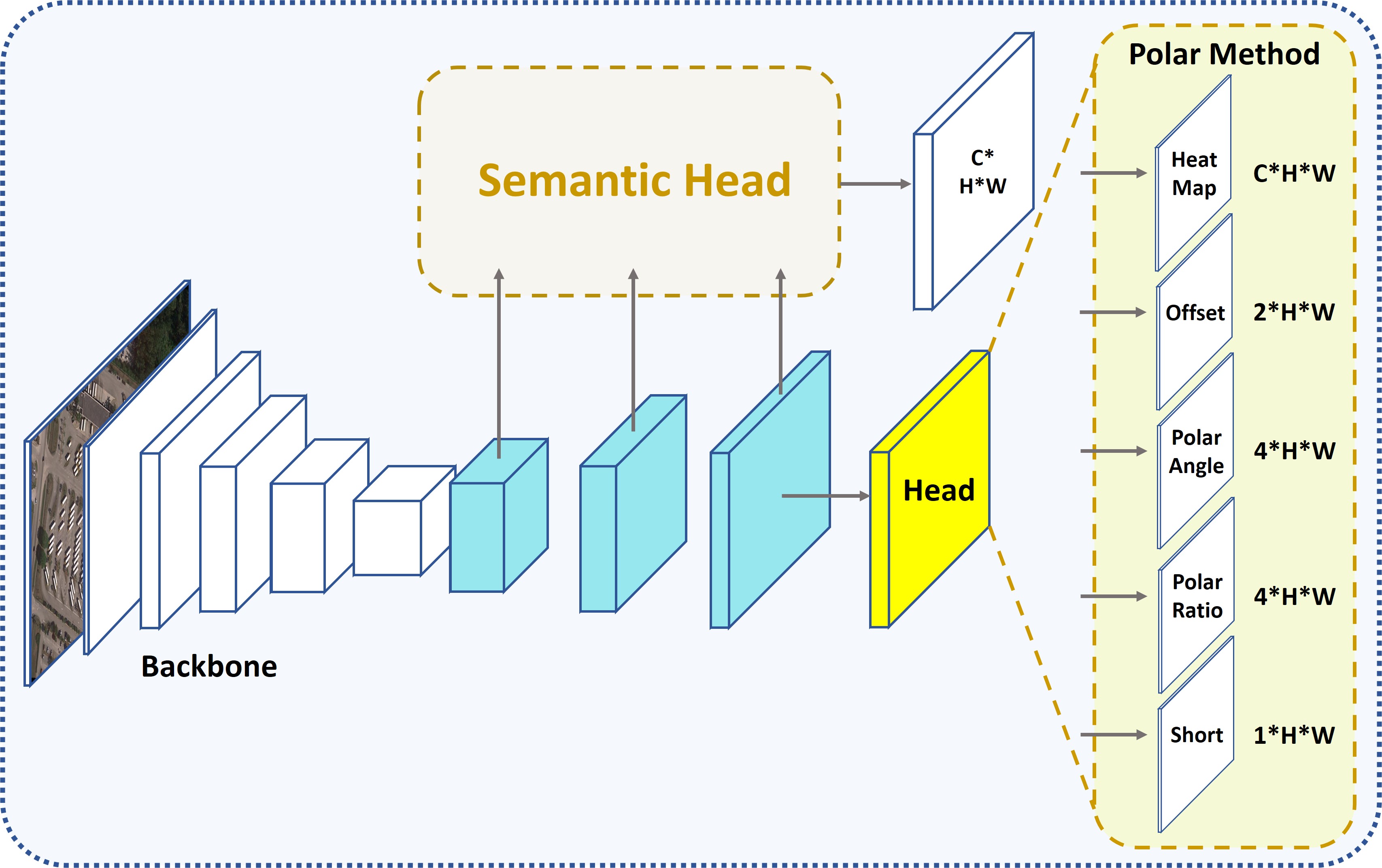}
	\caption{The pipeline of our proposed method. Our PolarDet mainly consists of five modules: basic backbone for feature extraction, 3x DeConv Modules for feature reconstruction, Polar method for regressing the size and angle of targets, Improved Center-Semantic Structure for decoding location and classification information which are hidden in sub-pixel.}
	\label{fig:structure}
\end{figure}

\subsection{Principle of Commonly used Representation}\label{PCR}

\begin{figure}[!tb]	
	\centering
	\includegraphics[width=1.0\linewidth]{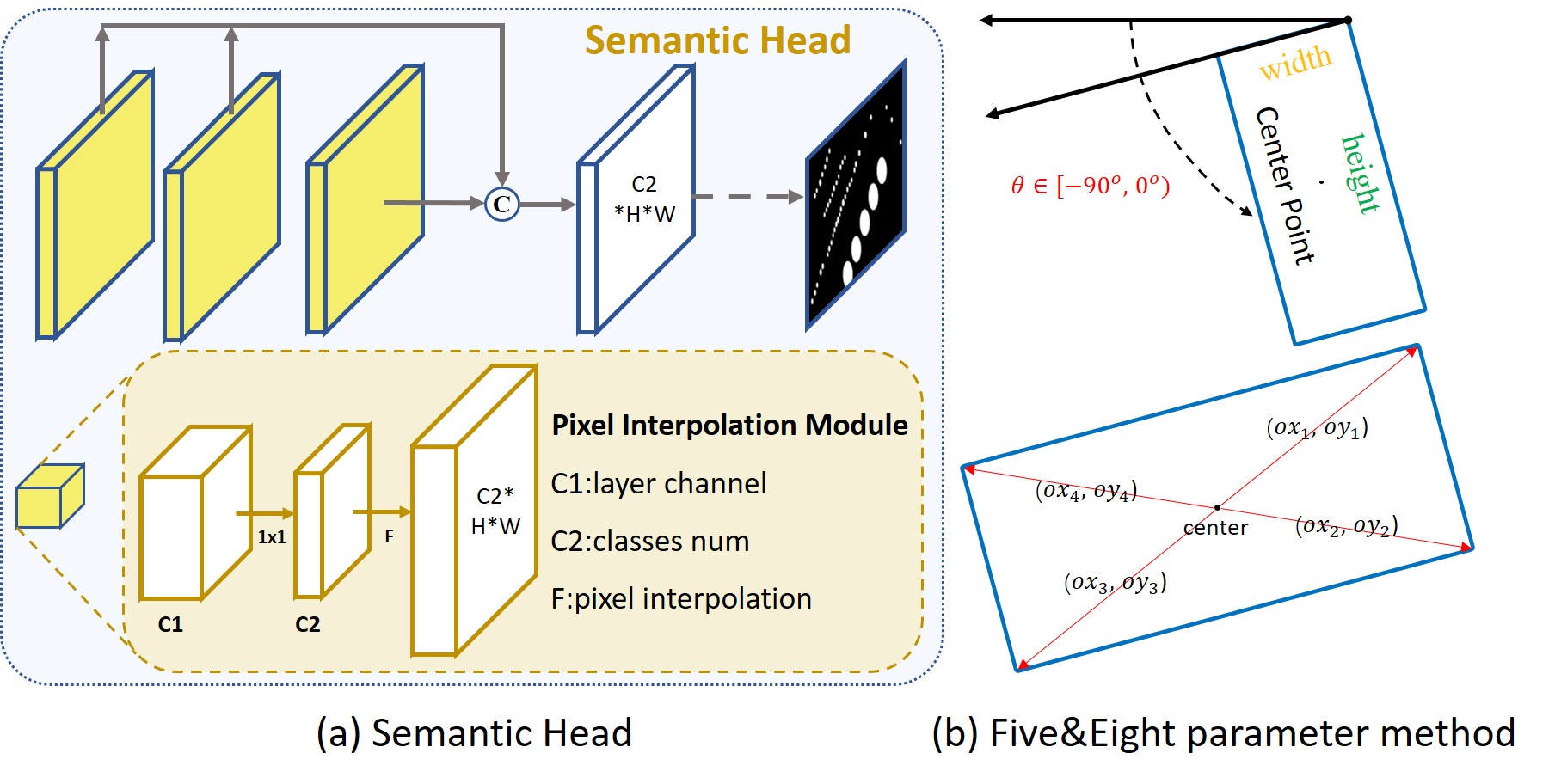}
	\caption{Left: Center-Semantic Head is composed of three Pixel Interpolation Module, where C1 means layer channel, C2 means class num, and F means pixel interpolation.\\Right: Five parameter and eight parameter methods for expressing targets}
	\label{fig:merge_new}
\end{figure}

In orientated detection tasks, nearly all the methods use five-parameter or eight-parameter to represent rotate targets. As shown in Fig.\ref{fig:merge_new}(b), the definition of five-parameter is based on OpenCV. (a) regard the point with the least y coordinate value as the vertex. (b) elicit a reference line from the vertex and coincide with the horizontal line. (c) rotate reference line counterclockwise until it covers the first side of the target, which is labeled as w(width) while the other side as h(height). (d) the center point is $(x,y)$, and orientation is equal to the rotation angle of the reference line.

As shown in Fig.\ref{fig:merge_new}(b), eight-parameter usually uses four boundary points to describe the rotating target. These four vertices are often represented as offsets from the center point. Also, they are defined in counterclockwise order where the original point is the one with the least $y$ coordinate value.

However, these two methods may meet problems such as angle boundary, angle-loss trap, and convergence performance decrease. These problems will cause the decrease in dectection precision.


\subsection{Pipeline}\label{structure}
As shown in Fig.\ref{fig:structure} and Fig.\ref{fig:merge_new}(a), PolarDet consists of the following five modules. The first is the feature extraction element, which is based on ResNet. In our method, we use ResNet18 in Ablation Study and ResNet101 in final testing. The following feature reconstruction increases the resolution to quarter than the input image. We use common deconvolution combined with the dcn module to expand the receptive field. From the last feature map, the polar method regresses five heads to represent the target. The regression contains Heatmaps, Offset, Polar-Angle, Shorter, and Polar-Ratio, which express the center of target, the offset of heatmap, angles between four polar diameters and the reference y-axis, the shorter one between the width and height of the target minimum bounding rectangle, the length ratio between shorter side and polar diameter, respectively. Also, we deploy the center semantic structure to enhance the accuracy of classification. This kind of light network design can maintain the inference by using interpolation.

\subsection{Center Point}\label{center point}

As discussed above, our method uses a center point to locate and classify targets. Different from CenterNet\cite{zhou2019objects}, we apply the center of minimum bounding rectangle for detecting rotate target or quadrilateral. But we still maintain the truth value of heatmap, which is actually a confidence map with value range from $h \in [0,1]^{C \times \frac{H}{4} \times \frac{W}{4}}$. $C$ is the number of dataset category.

In the training stage, we generate a truth-heatmap using the Gaussian kernel to train the confidence map. In detail, we first map targets onto a single point, which can be expressed as $H_{c,y,x}=1$. Then we use Gaussian kernel to endow value to the neighboring points followed by CornerNet, which can be expressed as ${H_{c,y,x} = \exp\left(-\frac{(x-\tilde p_x)^2+(y-\tilde p_y)^2}{2\sigma_p^2}\right)}$, where $\sigma_p$ follows the definition in CenterNet.  Finally, as shown in formula.\ref{equ1}, center focal loss \cite{lin2017focal} is applied to guide the direction of regression:
\begin{equation}\label{equ1}
L_k = \frac{-1}{N} \sum_{xyc}
\begin{cases}
(1 - \hat{H}_{c,y,x})^{\alpha} 
\log(\hat{H}_{c,y,x}) & \!\text{if}\ H_{c,y,x}=1\vspace{2mm}\\
\begin{array}{c}
(1-H_{c,y,x})^{\beta} 
(\hat{H}_{c,y,x})^{\alpha}\\
\log(1-\hat{H}_{c,y,x})
\end{array}
& \!\text{otherwise}
\end{cases}
\end{equation}
where N is the number of objects in input image, $\alpha$ and $\beta$ are hyper-parameters which are set to 2 and 4 respectively.

\subsection{Polar Angle}\label{polar_angle}
\emph{A. Definition}

\begin{figure}[!tb]	
	\centering
	\includegraphics[width=0.8\linewidth]{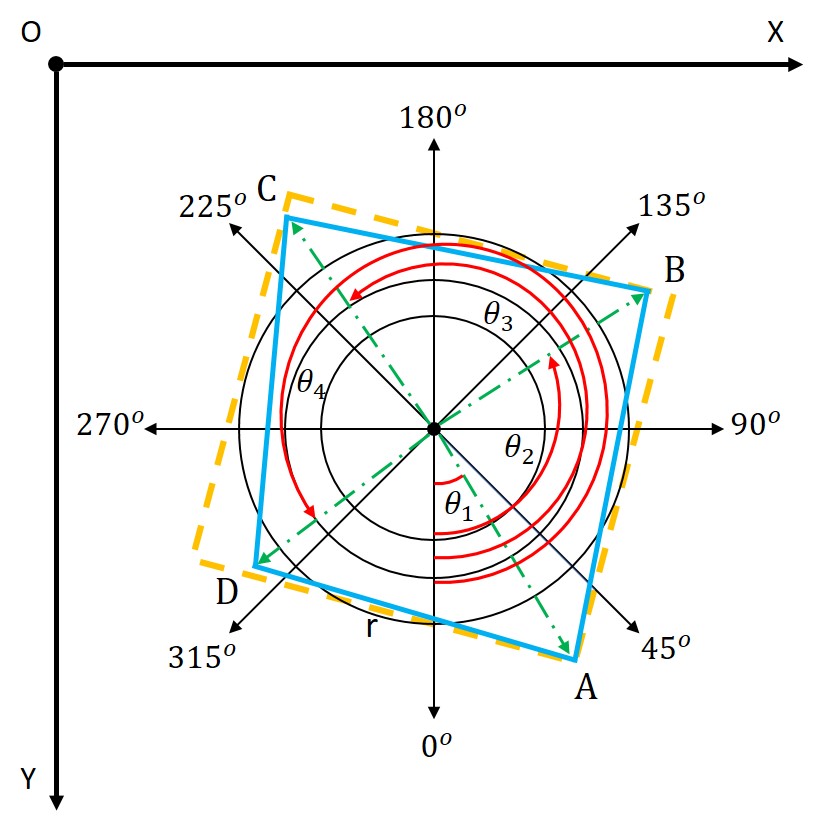}
	\caption{Polar Angle representation method. OX and OY means the positive axis of the image coordinate system, blue box means target, orange box means the minimum bounding rectangle, dotted lines direct the vertices of the target}
	\label{fig:polar method}
\end{figure}

As shown in Fig.\ref{fig:polar method}, angles are based on the polar coordinate system. The definition of the positive axis is along with the image positive x-axis and the positive y-axis respectively. We regard the beginning angle $0^o$ coincide with the positive y-axis and increase the angle counterclockwise. In our method, we use four angles to describe the orientation of targets, which defined as $(\theta_1,\theta_2,\theta_3,\theta_4)$. Making quadrilateral as an example, as the blue box shown in Fig.\ref{fig:polar method}, $A,B,C,D$ express the four points of the quadrilateral. Under our polar coordinate system, we define $(\theta_1,\theta_2,\theta_3,\theta_4)$ ranging from $0$ to $2\pi$ which follow the radian. Also, these four angles increase counterclockwise, which means $\theta_1$ will be the smallest angle while the $\theta_4$ will be the biggest one. In our Polar Angle definition method, the range of angles makes them express targets flexibly. Moreover, the radian guarantees the stability of regression. In the training stage, we simply use L1-Loss as regression loss (as shown in formula.\ref{equ2}).
\begin{equation}
L_{A} = \frac{1}{N}\sum_{m}^{i} \left|\hat A_{i} - A_i\right|
\label{equ2}
\end{equation}

\emph{B. Angle-Loss Trap Avoidance}

\begin{figure}[!tb]	
	\centering
	\includegraphics[width=0.9\linewidth]{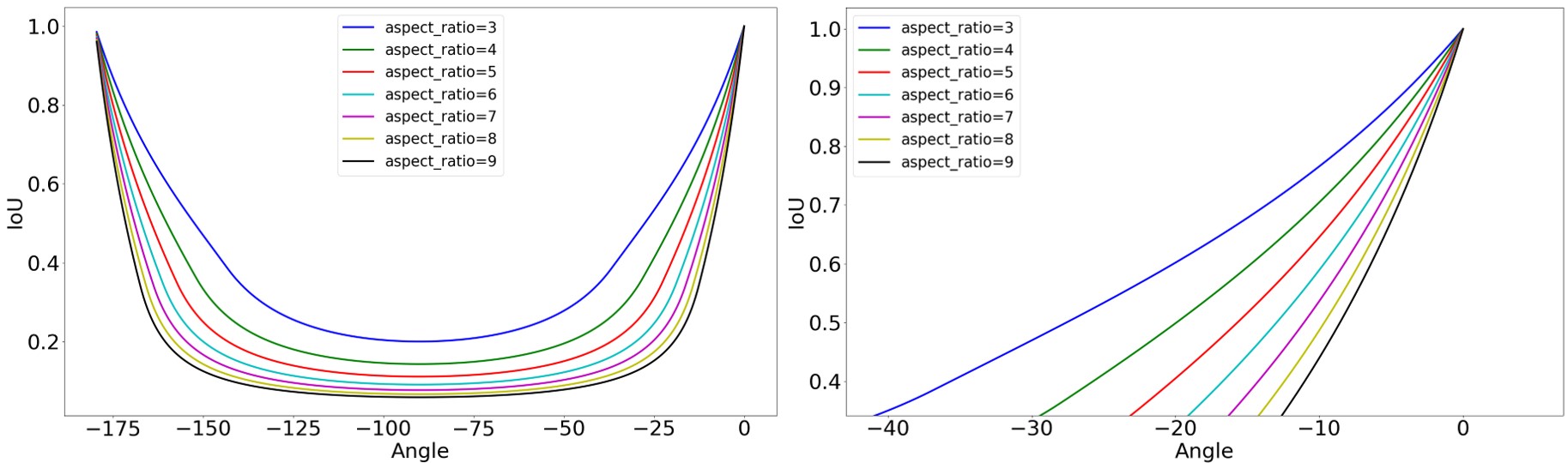}
	\caption{The relationship between one-angle bias and IOU with different aspect-ratio.}
	\label{fig:IOUandAngle}
\end{figure}

Many detectors that express the oriented target through just one angle may fall into the Angle-Loss Trap in a very high probability. As shown in \ref{PCR}, the angle of five-parameter method ranges from $-90^o$ to $0^o$. When this definition combined with radian strategy, the angle will only range from $-0.5\pi$ to $0$, which is a relatively small value. As shown in Fig.\ref{fig:resolve problem}(a)(b)(c), the IOU highly relies on the precision of the angle when the aspect-ratio is relatively high. Under this situation, as shown in Fig.\ref{fig:IOUandAngle}, the IOU will drop a lot even the angle just misses a little. Worse, when the angle misses a little, the angle loss will be still very small because of the value range of rotation angle, which means dropping into the Angle-Loss Trap. For avoiding this trap, we propose the polar angle for substituting. As shown in Fig.\ref{fig:resolve problem}(c), the polar angle method can express oriented target to the point. Also, the polar angle method can enhance angle convergence performance by the larger angle loss created by four angles expression.

\emph{C. Boundary Problem Solution}

\begin{figure}[!tb]	
	\centering
	\includegraphics[width=0.8\linewidth]{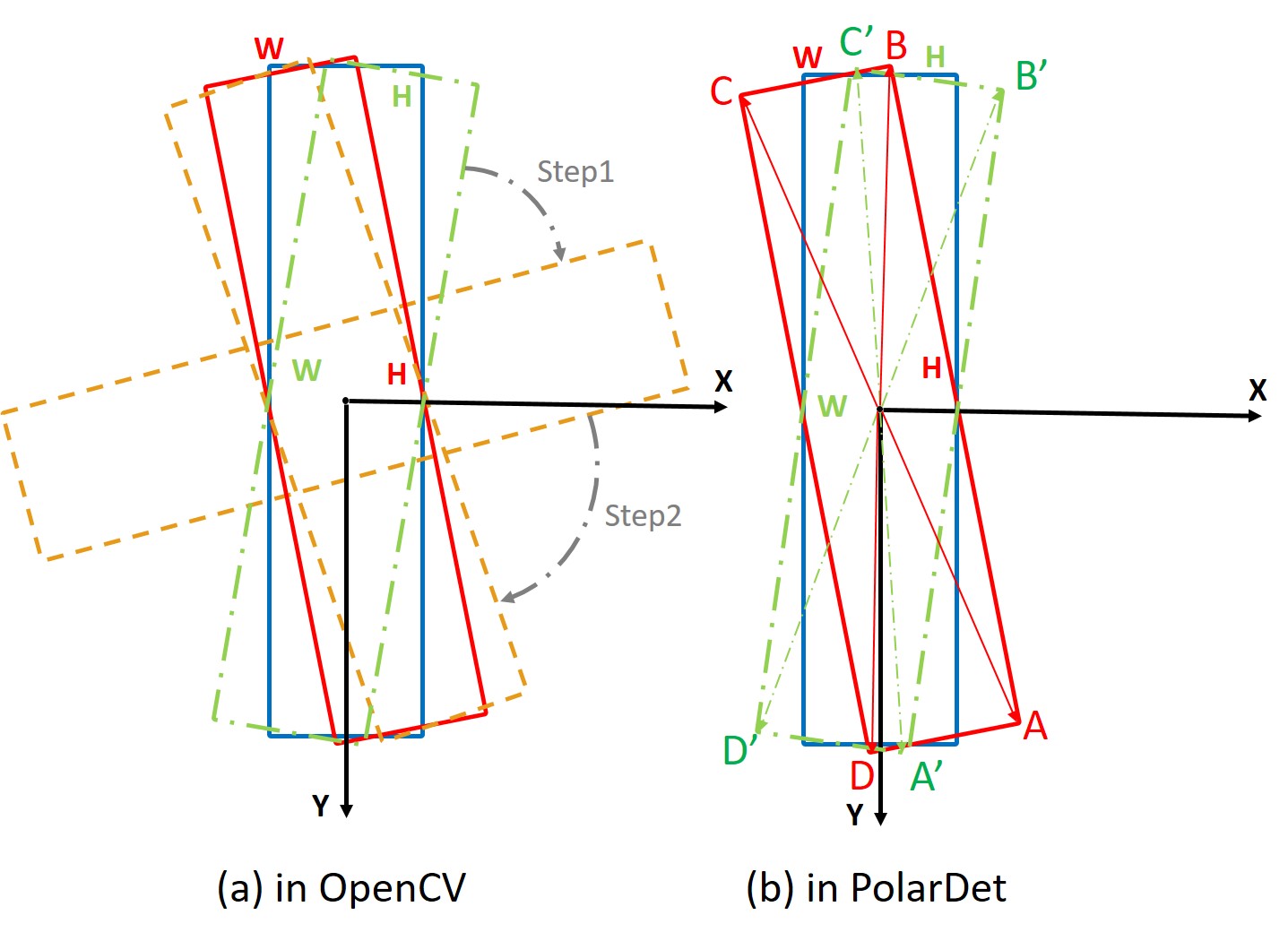}
	\caption{comparison in different angle definition: (a) prediction has to regress by step1 and step2 with OpenCV angle definition (b) regress directly using PolarDet angle definition}
	\label{fig:boundary problem}
\end{figure}

Boundary problem is always existing trouble when using the five-parameter expression. As shown in Fig.\ref{fig:boundary problem}(a), blue, green, red boxes represent the reference box, prediction box, gt box respectively. We can notice that the green box seems only to miss a little in angle prediction. However, the angle of gt is $-10^o$ and the angle of prediction is $-80^o$, which misses a lot. Worse, as the figure shows, the width and height of the target are opposite with ground-truth, which also creates a very big loss. This situation leads the network to have to learn in the following steps. (1) first, the network needs to make the angle decrease to $-10^o$ as shown in step1. (2) then the network has to force the width into a smaller one while the height into a bigger one. These two steps will make the network unstable and decrease convergence performance. Our PolarDet based on four Polar Angles can easily resolve this problem. As shown in Fig.\ref{fig:boundary problem}(b), all the points $A,B,C,D$ in ground-truth are correspond to $A^{'},B^{'},C^{'},D^{'}$ in prediction respectively. Because we use a more precise angle range strategy described in Fig.\ref{fig:polar method}. All the predictions only miss a little compared with ground-truth. Also, the length will be close to ground-truth according to the polar diameter expression (discussing in \ref{polar_diameter}). With these mentioned above, the PolarDet will converge to the truth quickly and won’t be bothered with the boundary problem.

\subsection{Polar Diameter}\label{polar_diameter}
\emph{A. Definition}

In PolarDet, instead of using commonly used width and height to describe the size of target, we introduce a polar diameter expression. We introduce one length parameter and a pair of ratio parameters $(s,r_1,r_2,r_3,r_4)$ to express target. In detail, as shown in Fig.\ref{fig:polar method}, the expression consists of the following two parts: (1) the shorter side of the minimum bounding rectangle of target. (2) four ratios between shorter side and polar diameters. The polar diameters mean the distance between vertices and center point shown in \ref{PCR}. In the training stage, we still use L1-Loss for both of the expressions. These definitions and loss are shown directly in formula \ref{equ3} \ref{equ4} \ref{equ5} \ref{equ6}, where $s$ means shorter side, $MBR$ means the minimum bounding rectangle of target, $r$ means ratio, $D_{L2}$ means $L2$ distance, $C$ means center point and $V_i$ means four vertices.
\begin{equation}
s = min(w_{MBR},h_{MBR})
\label{equ3}
\end{equation}
\begin{equation}
r = \frac{s}{D_{L2}(C,V_{i})} (i=1,2,3,4)
\label{equ4}
\end{equation}
\begin{equation}
L_{s} = \frac{1}{N}\sum_{m}^{i} \left|\hat s_{i} - s_i\right|
\label{equ5}
\end{equation}
\begin{equation}
L_{r} = \frac{1}{N}\sum_{4}^{k}\sum_{m}^{i}\left|\hat r_{ik} - r_{ik}\right|
\label{equ6}
\end{equation}

\begin{figure}[!tb]	
	\centering
	\includegraphics[width=0.9\linewidth]{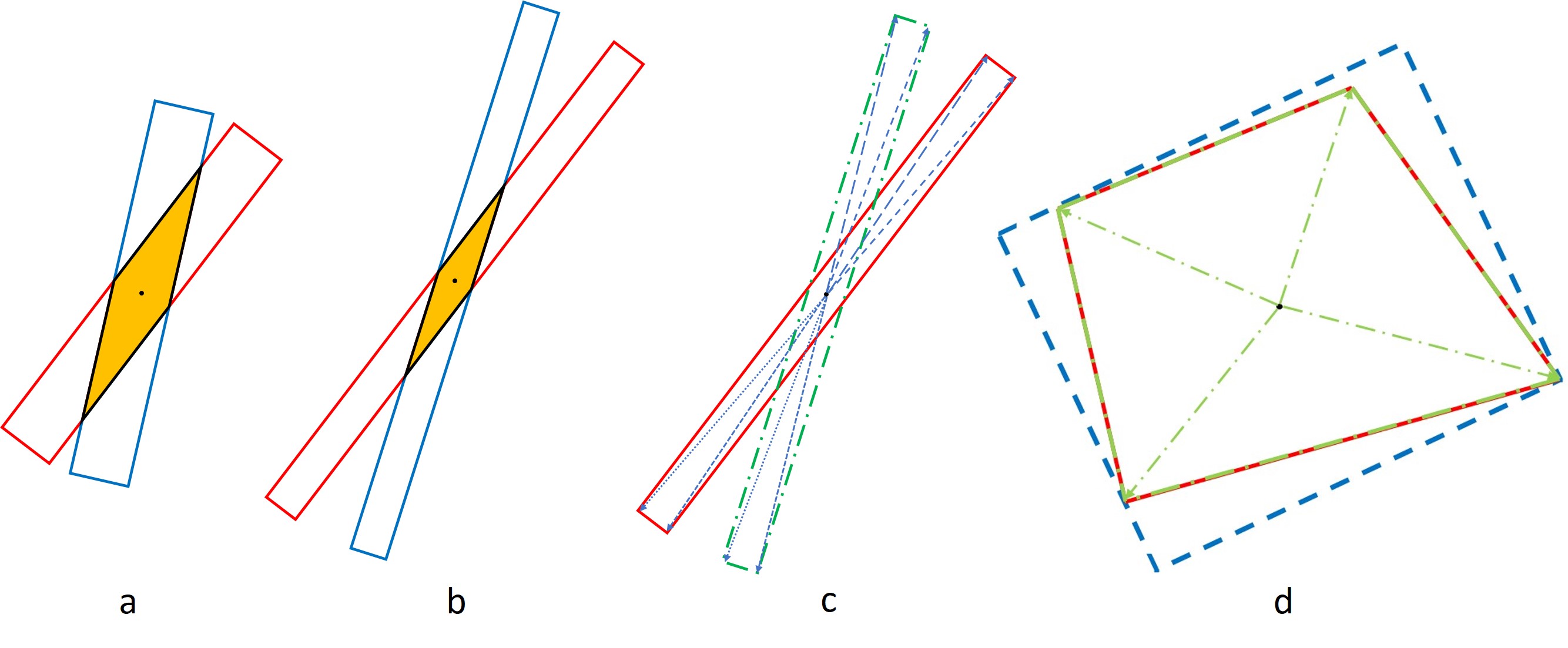}
	\caption{(a) and (b) show the IOU dropping a lot even with a small one-angle bias. (c) shows the superiority of our Angle Polar representation method. (d) shows different expression for quadrilateral or orientation.}
	\label{fig:resolve problem}
\end{figure}

\begin{table*}[!tb]
	\centering
	\resizebox{0.98\textwidth}{!}{
		\begin{tabular}{l|c|c|c|c|c|c|c|c|c|c|c|c|c|c|c|c|c|c}
			\hline
			Backbone & System & Expression & PL &  BD &  BR &  GTF &  SV &  LV &  SH &  TC &  BC &  ST &  SBF &  RA &  HA &  SP &  HC &  mAP \\
			\hline
			\multirow{5}{*}{ResNet-18}
			& Cartesian & Single Angle & 89.79 & 71.73 & 25.27 & 52.93 & 51.59 & 75.54 & 78.00 & 90.75 & 63.77 & 78.19 & 61.48 & 67.08 & 61.04 & 69.28 & 42.99 & 65.29 \\
			& Polar & Direct & 89.98 & 67.69 & 29.66 & 59.38 & 61.58 & 75.39 & 77.25 & 90.77 & 64.76 & 79.30 & 63.08 & 64.39 & 63.35 & 59.96 & 46.59 & 66.21 \\
			& Polar & Average & 89.65 & 75.88 & 27.89 & 53.14 & 60.37 & 74.25 & 77.05 & 90.73 & 63.42 & 78.35 & 64.40 & 64.44 & 62.95 & 60.62 & 44.76 & 65.86 \\ 
			& Polar & Longer+Ratio & 89.81 & 66.54 & 29.33 & 54.04 & 61.24 & 74.75 & 77.33 & 90.72 & 59.60 & 78.76 & 62.46 & 62.77 & 62.48 & 59.54 & 42.70 & 64.80 \\
			& Polar & Shorter+Ratio & 89.93 & 76.56 & 34.65 & 59.84 & 67.06 & 78.28 & 86.12 & 90.81 & 66.13 & 80.10 & 70.58 & 60.72 & 64.64 & 60.55 & 53.37 & \bf{69.29} \\
			\hline
	\end{tabular}}
	\caption{AP and mAP (\%) about using different kinds of target expression method}
	\label{ablation study of expression}
\end{table*}
\begin{table*}[!tb]
	\centering
	\resizebox{0.98\textwidth}{!}{
		\begin{tabular}{l|c|c|c|c|c|c|c|c|c|c|c|c|c|c|c|c|c|c}
			\hline
			Backbone & Polar-Method & Center-Semantic & PL &  BD &  BR &  GTF &  SV &  LV &  SH &  TC &  BC &  ST &  SBF &  RA &  HA &  SP &  HC &  mAP \\
			\hline
			\multirow{3}{*}{ResNet-18}
			& - & - & 89.79 & 71.73 & 25.27 & 52.93 & 51.59 & 75.54 & 78.00 & 90.75 & 63.77 & 78.19 & 61.48 & 67.08 & 61.04 & 69.28 & 42.99 & 65.29 \\
			& $\surd$ & - & 89.93 & 76.56 & 34.65 & 59.84 & 67.06 & 78.28 & 86.12 & 90.81 & 66.13 & 80.10 & 70.58 & 60.72 & 64.64 & 60.55 & 53.37 & 69.29 \\
			& $\surd$ & $\surd$ & 89.97 & 75.64 & 33.54 & 54.61 & 66.11 & 78.12 & 85.65 & 90.84 & 67.72 & 78.80 & 72.29 & 64.32 & 65.41 & 59.71 & 68.00 & \bf{70.05} \\
			\hline
	\end{tabular}}
	\caption{AP and mAP (\%) about applying different strategies of classification enhancement or not applying}
	\label{ablation study of Center-Semantic}
\end{table*}

\emph{B. Convergence Performance Increase}\label{RCS}

Whatever angle expression the common methods use, they often apply width and height to describe the size of the target. However, we notice that this kind of length expression often meets the decrease in network convergence performance. This decrease is commonly caused by sharp changes in the length regression, especially for tasks with various scales targets. For raising the performance of convergence, we propose the relative polar diameter expression which uses the shorter side and ratio discussed above. This strategy will decrease the prediction range and increase performance. As shown in Table \ref{ablation study of expression}, our expression has a better performance than the other methods including the ones using the Cartesian system and the ones using other expressions with the Polar system. These results all regard ResNet-18 as backbone and in the same hyper-parameter.
\\

\emph{C. More Precise Quadrilateral Fitting}

In oriented object detection like the aerial field, labels are often given as four detached points. Therefore, the ground-truth often manifests as quadrilateral.

As shown in Fig.\ref{fig:resolve problem}(d), the red, blue, green box represents ground-truth, minimum bounding rectangle, polar angle with polar diameter expression. Easily, we can notice that just use minimum bounding rectangle (MBR) cannot cover target completely. Background information will also be included in MBR, which will confuse the network. However, when applying our method into expression, quadrilateral can be totally covered without introducing useless information. We will show our impressive results in detecting orientated targets in the following experiments in section \ref{experiments}.

\subsection{Improved Center-Semantic Structure}\label{Center_Semantic}
Clutter and complex background in aerial images will decrease classification performance and cause false positive detection.  Inspired by PointRend\cite{kirillov2020pointrend}, which introduces boundary information exists in sub-pixel, we propose that classification information also exists in sub-pixel. Therefore, our method introduces an improved center-semantic structure to optimize classification and verify our idea. As shown in Fig.\ref{fig:merge_new}(a), we generate three pixel-interpolation modules $P^{C \times \frac{H}{4} \times \frac{W}{4}}$, then we concat them into merge layer $M^{C \times \frac{H}{4} \times \frac{W}{4}}$ and multiple it with the predicted heatmap pixel-wisely in inference period.

In detail, we use one $1\times1$ conv and four times bilinear upsample, one $1\times1$ conv and two times bilinear upsample, only one $1\times1$ conv to get the first, second, third pixel-interpolation modules. Then, we concat these three modules into $M^{'C \times \frac{H}{4} \times \frac{W}{4}}$. Finally, we use one $3\times3$ conv followed by $1\times1$ conv to get merge layer $M^{C \times \frac{H}{4} \times \frac{W}{4}}$, where each ground-truth will be expressed as a circle with the same diameter as in the confidence map. However, the value of the foreground will all be 1 while the background will be 0, which is different from the Gaussian value. In other words, by category classification heatmap $hc \in [0,1]^{C \times \frac{H}{4} \times \frac{W}{4}}$. 


In this structure, we apply two strategies to reduce the parameter. First, we use $1\times1$ conv to change the number of channels into the number of class. Then, we introduce bilinear interpolation but not the commonly used deconvolution to upsample the resolution. With these, we can increase classification performance without increasing much parameter. Also, the following Table \ref{ablation study of Center-Semantic} shows that our center-semantic can get obvious improvement compared with baseline.

\section{Experiments}\label{experiments}

\subsection{Datasets}

We choose a wide range of different type datasets containing plentiful oriented targets, which are taken by satellite, drone, helicopter. The details are as follows.

\emph{A. DOTA}\label{DOTA}

DOTA \cite{xia2018dota} is the largest dataset for oriented object detection which contains 2,806 aerial images and 15 categories with almost 200,000 instances. In the DOTA dataset, the training set, validation set, and test set account for 1/2, 1/6, 1/3 of the whole set, respectively. 

We use the training set for training and validation set for evaluation in the ablation study while both the training and validation set for training and test set for submitting in the final test. Because the resolution of DOTA ranges from 800*800 to 30000*30000, we split the image into 1024*1024 patches with an overlap of 200 pixels, which is the same as the others like ROI Transformer \cite{ding2019learning}. With this, we get about 14,000 patches and 19,000 patches in double tasks mentioned above. The model is trained by 360 epochs, and the learning rate drops from 1.25e-4 to 1.25e-6 in the 200$^{th}$ epoch and 300$^{th}$ epoch.

\emph{B. UCAS-AOD}\label{UCAS_AOD}

UCAS-AOD \cite{li2019feature} is a specialized dataset for remote sensing target detection. It contains 1,510 images with about 15,000 instances in two categories including plane and car. In line with SCRDet++ \cite{yang2020scrdet++}, we randomly select 1,110 for training and 400 for testing.

\begin{table*}[!tb]
	\centering
	\resizebox{0.98\textwidth}{!}{
		\begin{tabular}{l|c|c|c|c|c|c|c|c|c|c|c|c|c|c|c|c|c}
			\hline
			\textbf{OBB (oriented bounding boxes)} & Backbone &  PL &  BD &  BR &  GTF &  SV &  LV &  SH &  TC &  BC &  ST &  SBF &  RA &  HA &  SP &  HC &  mAP\\
			\hline
			\textbf{Two-Stage Methods} & \multicolumn{16}{|c}{} \\
			\hline
			FR-O \cite{xia2018dota} & ResNet101 \cite{he2016deep} & 79.09 & 69.12 & 17.17 & 63.49 & 34.20 & 37.16 & 36.20 & 89.19 & 69.60 & 58.96 & 49.4 & 52.52 & 46.69 & 44.80 & 46.30 & 52.93 \\
			R$^2$CNN \cite{jiang2017r2cnn} & ResNet101 & 80.94 & 65.67 & 35.34 & 67.44 & 59.92 & 50.91 & 55.81 & 90.67 & 66.92 & 72.39 & 55.06 & 52.23 & 55.14 & 53.35 & 48.22 & 60.67 \\
			RRPN \cite{ma2018arbitrary} & ResNet101 & 88.52 & 71.20 & 31.66 & 59.30 & 51.85 & 56.19 & 57.25 & 90.81 & 72.84 & 67.38 & 56.69 & 52.84 & 53.08 & 51.94 & 53.58 & 61.01 \\
			ICN \cite{azimi2018towards} & ResNet101 & 81.40 & 74.30 & 47.70 & 70.30 & 64.90 & 67.80 & 70.00 & 90.80 & 79.10 & 78.20 & 53.60 & 62.90 & 67.00 & 64.20 & 50.20 & 68.20 \\
			RADet \cite{li2020radet} & ResNeXt101 \cite{xie2017aggregated} & 79.45 & 76.99 & 48.05 & 65.83 & 65.46 & 74.40 & 68.86 & 89.70 & 78.14 & 74.97 & 49.92 & 64.63 & 66.14 & 71.58 & 62.16 & 69.09 \\
			RoI-Transformer \cite{ding2019learning} & ResNet101 & 88.64 & 78.52 & 43.44 & 75.92 & 68.81 & 73.68 & 83.59 & 90.74 & 77.27 & 81.46 & 58.39 & 53.54 & 62.83 & 58.93 & 47.67 & 69.56 \\
			CAD-Net \cite{zhang2019cad} & ResNet101 & 87.8 & 82.4 & 49.4 & 73.5 & 71.1 & 63.5 & 76.7 & 90.9 & 79.2 & 73.3 & 48.4 & 60.9 & 62.0 & 67.0 & 62.2 & 69.9 \\
			SCRDet \cite{yang2019scrdet} & ResNet101 & 89.98 & 80.65 & 52.09 & 68.36 & 68.36 & 60.32 & 72.41 & 90.85 & 87.94 & 86.86 & 65.02 & 66.68 & 66.25 & 68.24 & 65.21 & 72.61\\	
			FADet \cite{li2019feature} & ResNet101 & 90.21 & 79.58 & 45.49 & 76.41 & 73.18 & 68.27 & 79.56 & 90.83 & 83.40 & 84.68 & 53.40 & 65.42 & 74.17 & 69.69 & 64.86 & 73.28\\
			Gliding Vertex \cite{xu2020gliding} & ResNet101 & 89.64 & 85.00 & 52.26 & 77.34 & 73.01 & 73.14 & 86.82 & 90.74 & 79.02 & 86.81 & 59.55 & 70.91 & 72.94 & 70.86 & 57.32 & 75.02 \\
			Mask OBB \cite{wang2019mask} & ResNeXt101 & 89.56 & 85.95 & 54.21 & 72.90 & 76.52 & 74.16 & 85.63 & 89.85 & 83.81 & 86.48 & 54.89 & 69.64 & 73.94 & 69.06 & 63.32 & 75.33 \\
			FFA \cite{fu2020rotation} & ResNet101 & 90.1 & 82.7 & 54.2 & 75.2 & 71.0 & 79.9 & 83.5 & 90.7 & 83.9 & 84.6 & 61.2 & 68.0 & 70.7 & 76.0 & 63.7 & 75.7 \\
			APE \cite{zhu2020adaptive} & ResNeXt-101 & 89.96 & 83.62 & 53.42 & 76.03 & 74.01 & 77.16 & 79.45 & 90.83 & 87.15 & 84.51 & 67.72 & 60.33 & 74.61 & 71.84 & 65.55 & 75.75 \\
			SCRDet++-MS \cite{yang2020scrdet++} & ResNet101 & 90.05 & 84.39 & 55.44 & 73.99 & 77.54 & 71.11 & 86.05 & 90.67 & 87.32 & 87.08 & 69.62 & 68.90 & 73.74 & 71.29 & 65.08 & 76.81 \\
			\hline
			\textbf{Single-Stage Methods} & \multicolumn{16}{|c}{} \\
			\hline
			IENet \cite{lin2019ienet} & ResNet101 & 80.20 & 64.54 & 39.82 & 32.07 & 49.71 & 65.01 & 52.58 & 81.45 & 44.66 & 78.51 & 46.54 & 56.73 & 64.40 & 64.24 & 36.75 & 57.14 \\
			RetinaNet \cite{lin2017focal} & ResNet101 & 88.92 & 67.67 & 33.55 & 56.83 & 66.11 & 73.28 & 75.24 & 90.87 & 73.95 & 75.07 & 43.77 & 56.72 & 51.05 & 55.86 & 21.46 & 62.02 \\
			P-RSDet \cite{zhou2020objects} & ResNet101 & 89.02 & 73.65 & 47.33 & \textbf{72.03} & 70.58 & 73.71 & 72.76 & 90.82 & 80.12 & 81.32 & 59.45 & 57.87 & 60.79 & 65.21 & 52.59 & 69.82 \\
			O$^2$-DNet \cite{wei2019oriented} & Hourglass104 \cite{newell2016stacked} & 89.31 & 82.14 & 47.33 & 61.21 & 71.32 & 74.03 & 78.62 & 90.76 & 82.23 & 81.36 & 60.93 & 60.17 & 58.21 & 66.98 & 61.03 & 71.04 \\
			R$^3$Det \cite{yang2019r3det} & ResNet152 & 89.24 & 80.81 & 51.11 & 65.62 & 70.67 & 76.03 & 78.32 & 90.83 & 84.89 & 84.42 & 65.10 & 57.18 & 68.10 & 68.98 & 60.88 & 72.81\\
			RSDet \cite{qian2019learning} & ResNet152 & \textbf{90.1} & 82.0 & \textbf{53.8} & 68.5 & 70.2 & 78.7 & 73.6 & \textbf{91.2} & \textbf{87.1} & 84.7 & 64.3 & 68.2 & 66.1 & 69.3 & 63.7 & 74.1 \\ 
			SCRDet++ \cite{yang2020scrdet++} & ResNet152 & 89.20 & 83.36 & 50.92 & 68.17 & 71.61 & 80.23 & 78.53 & 90.83 & 86.09 & 84.04 & \textbf{65.93} & 60.8 & 68.83 & 71.31 & 66.24 & 74.41 \\
			PolarDet (ours) & ResNet50 & 89.73 & \textbf{87.05} & 45.30 & 63.32 & \textbf{78.44} & 76.65 & \textbf{87.13} & 90.79 & 80.58 & \textbf{85.89} & 60.97 & 67.94 & \textbf{68.20} & \textbf{74.63} & \textbf{68.67} & \textbf{75.02} \\
			PolarDet-MS (ours) & ResNet101 & 89.65 & \textbf{87.07} & 48.14 & 70.97 & \textbf{78.53} & \textbf{80.34} & \textbf{87.45} & 90.76 & 85.63 & \textbf{86.87} & 61.64 & \textbf{70.32} & \textbf{71.92} & \textbf{73.09} & \textbf{67.15} & \textbf{76.64} \\
			\hline
	\end{tabular}}
	\caption{AP and mAP (\%) across categories of OBB task on DOTA.\\FT indicates horizontal-flip MS indicates multi-scale}
	\label{DOTA result}
\end{table*}

\emph{C. HRSC2016}\label{HRSC2016}

HRSC2016 \cite{liu2016ship} is another challenging dataset in the aerial field for oriented target detection. It contains 1,061 images and more than 20 categories of ships in various appearances. The image resolution is about 1500*900 not many ships in one image. Following ROI Transformer, we use the trainval set (617 images) for training and the test set (444 images) for testing, and we resize the image into both 1024*1024 and 800*800 resolution in the same method used in the DOTA dataset. We only use level one to execute detection task like the other work such as ROI Transformer \cite{ding2019learning} and P-RSDet \cite{zhou2020objects}.

\subsection{Implementation Details}\label{Implementation_Details}
The experiments on DOTA, UCAS-AOD, and HRSC2016 are implemented by Pytorch 1.0. We use 2×32GB NVIDIA Tesla V100 GPUs in the ablation study and 8$\times$32GB NVIDIA Tesla V100 GPUs in final testing. We adopt ResNet-18 as backbone in all Ablation Studies and ResNet-101 in final testing. We set learning rate to 1.25e-4 then drop it by 10 times in 200$^{th}$, 300$^{th}$ epoch in DOTA, UCAS-AOD, HRSC2016, and ICDAR2015 datasets. For all the datasets, the network is trained by Adam \cite{kingma2014adam} optimizer with 128 batch size.

\begin{figure*}[!tb]
	\begin{center}
		\includegraphics[width=1.0\linewidth]{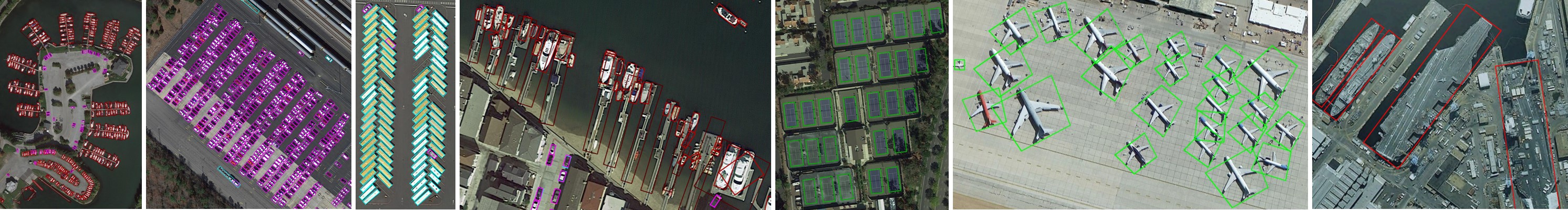}
	\end{center}
	\caption{Merged Illustration of our proposed method PolarDet. We achieve amazing consequence when detecting Dense, Oriented, clutter, and high aspect-ratio targets in different datasets including DOTA, UCAS-AOD, and HRSC2016.}
	\label{fig:show pics}
\end{figure*}

\subsection{Ablation Study}\label{Ablation_Study}
\begin{table}[tb!]
	\centering
	\resizebox{0.35\textwidth}{!}{
		\begin{tabular}{l|c|c|c}
			\hline
			Method & Plane & Car & mAP \\
			\hline
			YOLOv2 \cite{redmon2017yolo} & 96.60 & 79.20 & 87.90 \\
			R-DFPN \cite{yang2018automatic} & 95.90 & 82.50 & 89.20 \\
			DRBox \cite{liu2017learning} & 94.90 & 85.00 & 89.95 \\ 
			S$^2$ARN \cite{bao2019single} & 97.60 & 92.20 & 94.90 \\
			RetinaNet-H \cite{yang2019r3det} & 97.34 & 93.60 & 95.47 \\
			ICN \cite{azimi2018towards} & -- & -- & 95.67 \\
			FADet \cite{li2019feature} & 98.69 & 92.72 & 95.71 \\
			R$^3$Det \cite{yang2019r3det} & 98.20 & 94.14 & 96.17 \\
			SCRDet++ \cite{yang2020scrdet++} & 98.93 & \textbf{94.97} & 96.95 \\
			\hline
			PolarDet (Ours) & \textbf{99.08} & {94.96} & \textbf{97.02} \\
			\hline
	\end{tabular}}
	\caption{Results by mAP (\%) on UCAS-AOD dataset.}
	\label{UCAS-AOD result}
\end{table}
\begin{table}
	\centering
	\resizebox{0.48\textwidth}{!}{
		\begin{tabular}{l|c|c|c|c|c}
			\hline
			Method & Backbone &  Image Size & Data Aug. &  mAP &  Speed \\
			\hline
			R$^2$CNN \cite{jiang2017r2cnn} & ResNet101 & 800*800 & $\times$ & 73.07 & 2fps \\
			RC1 \& RC2 \cite{lb2017high} & VGG16 & -- & -- & 75.7 & $<$1fps \\
			RRPN \cite{ma2018arbitrary} & ResNet101 & 800*800 & $\times$ & 79.08 & 3.5fps \\
			R$^2$PN \cite{zhang2018toward} & VGG16 & -- & $\surd$ & 79.6 & $<$1fps \\
			RetinaNet-H \cite{yang2019r3det} & ResNet101 & 800*800 & $\surd$ & 82.89 & 14fps \\
			RRD \cite{liao2018rotation} & VGG16 & 384*384 & -- & 84.3 & slow \\
			RetinaNet-R & ResNet101 & 800*800 & $\surd$ & 89.18
			& 10fps \\
			RoI-Transformer \cite{ding2019learning} & ResNet101 & 512*800 & $\times$ & 86.20 & 6fps \\
			R$^3$Det \cite{yang2019r3det} & ResNet152 & 800*800 & $\surd$ & {89.33} & 10fps \\
			\hline
			\multirow{2}{*}{PolarDet (Ours)}
			& ResNet50 & 800*800 & $\surd$ & \bf{90.13} & \bf32fps \\
			& ResNet50 & 1024*1024 & $\surd$ & \bf{90.46} & \bf25fps \\
			\hline
	\end{tabular}}
	\caption{Results and speed comparison on HRSC2016.}
	\label{HRSC2016 result}
\end{table}

\textbf{Effect of Polar method.} The Polar method involves two parts of target expression strategies, which are polar angle and polar diameter. We design different comparative experiments to prove the advancement of the Polar method as shown in Table \ref{ablation study of expression}. We compare our Polar method with methods using different coordinate systems and different expression strategies. From the results, we can notice that directly express by polar angles and polar diameters can make a sense, but a little (about 0.34\% increase). Also, we can see even adapt ratio as an expression bridge, the long side cannot achieve good performance but will hurt the stability of network (about 1.06\% decrease), which is caused by sharp changes in targets side length. For example, if directly regress length, the prediction will range from $[4,300]$ in DOTA dataset. However, as shown in Fig.\ref{fig:show pics}, our method can handle the situation even with the high aspect-ratio or very clutter environment. With our PolarDet, the prediction range will drop to $(0,\sqrt{2}]$, which is beneficial to regression.  Moreover, we can notice from the Table \ref{ablation study of expression} that our method nearly achieves big improvement in every category. Especially in BR, SV, LV, SH, SBF, and HC, we get 5\%, 6\%, 4\%, 9\%, 7\%, 8\% increase respectively, which prove the Polar method can resolve the sharp changes problems.

\textbf{Effect of Center-Semantic.} Center-Semantic is a brand-new structure used to enhance the accuracy of classification. In our center-semantic, as shown in Fig.\ref{fig:merge_new}(a), we use pixel interpolation strategy to upsample and lightweight feature extraction to decode classification information. With these two features, center-semantic can achieve enhancement without adding many parameters. As shown in Table \ref{ablation study of Center-Semantic}, we compare the simple baseline, Polar method, and Polar method + Center-Semantic together to show the efficiency of center-semantic. Obviously, center-semantic gets increas compared with baseline and only the Polar method.

\subsection{Comparison with the SOTA Methods}\label{comp}

Our PolarDet is applied to DOTA, AOD, and HRSC datasets respectively.

\textbf{Results on DOTA.} As illustrated in Table \ref{DOTA result}, we compare our results with the other state-of-the-art results on DOTA. The results of the DOTA shown here are obtained from the DOTA evaluation server. Model result is trained on the trainval set and evaluated on the test set. We only execute the OBB task, which is also DOTA official recommends. Our single-stage PolarDet finally achieves the best performance 76.64\% with ResNet-101 and MS testing. This result has already exceeded all one-stage methods and equaled the SOTA two-stage methods. And when paying attention to inference speed, our PolarDet is much faster than the other one-stage methods with ResNet-50 backbone, which only consumes 40ms in inference. Most importantly, PolarDet reaches 75.02\% mAP without any testing strategy in the mentioned condition. Fig.\ref{fig:show pics} shows the aerial images of clutter, complex, and huge scenes.


\textbf{Results on UCAS-AOD.} We also get the best performance on UCAS-AOD, where we get 97.02\% with ResNet-50 backbone. Table \ref{UCAS-AOD result} shows the competitive results.

\textbf{Results on HRSC2016.} We see HRSC2016 as a ship dataset regardless of Fine-grained image categorization. As shown in Table \ref{HRSC2016 result}, we achieve 90.13\% mAP and 90.46\% mAP on 800*800 and 1024*1024 resolution respectively, which are both SOTA results. Moreover, PolarDet is also the fastest detector with 25fps.

\section{Conclusion}
In this paper, we propose a polar coordinate-based one-stage detector, PolarDet. PolarDet is a faster and more precise detector for oriented object detection. In our PolarDet, we propose polar method and center semantic structure to represent the target. The polar method redesigns the regression method based on polar angle and polar diameter expression. Furthermore, the center semantic structure enhances the accuracy of classification and location. All these methods gain a great improvement in performance. Extensive experiments on DOTA, UCAS-AOD, and HRSC2016  verify our approaches, where we achieve SOTA performance even compared with SOTA two-stage detectors.

Besides the oriented tasks and quadrilateral detection, our PolarDet can also detect polygon, concavity, and key-point, where PolarDet just needs to add extra detection points.

\section{Acknowledgement}

The research was supported by Hikvision Research Institute.

\bibliographystyle{abbrv}
\bibliography{PolarDet}

\end{document}